\documentclass[letterpaper, 10 pt, conference]{ieeeconf}  

\IEEEoverridecommandlockouts                              

\usepackage{graphics} 
\usepackage{epsfig} 
\usepackage{times} 
\usepackage{amsmath} 
\usepackage{amssymb}  
\usepackage[T1]{fontenc}  
\usepackage{bm}
\usepackage{url}
\usepackage{float}
\usepackage{multirow}
\usepackage{dblfloatfix} 
\title{\LARGE \bf
Scalable Radar-based Roadside Perception: Self-localization and Occupancy Heat Map for Traffic Analysis}

\author{Longfei Han$^{1,2}$, Qiuyu Xu$^{1}$, Klaus Kefferp\"utz$^{1,3}$,  Ying Lu$^{4}$, Gordon Elger$^{1,3}$ and J\"urgen Beyerer$^{2,5}$
\thanks{$^{1}$Application Center \guillemotright Connected Mobility and Infrastructure\guillemotleft , Fraunhofer IVI, Ingolstadt, Germany}
\thanks{$^{2}$Vision and Fusion Laboratory, Karlsruhe Institute of Technology (KIT), Karlsruhe, Germany}
\thanks{$^{3}$Technische Hochschule Ingolstadt, Ingolstadt, Germany}
\thanks{$^{4}$Technical University of Munich, Munich, Germany}
\thanks{$^{5}$Fraunhofer IOSB, Karlsruhe, Germany}
\thanks{Corresponding author: {\tt\small longfei.han@ivi.fraunhofer.de}}
}

\begin{document}

\maketitle
\thispagestyle{empty}
\pagestyle{empty}

\begin{abstract}

4D mmWave radar sensors are suitable for roadside perception in city-scale Intelligent Transportation Systems (ITS) due to their long sensing range, weatherproof functionality, simple mechanical design, and low manufacturing cost.
In this work, we investigate radar-based ITS for scalable traffic analysis. 
Localization of these radar sensors at city scale is a fundamental task in ITS.
For flexible sensor setups, it requires even more effort.  
To address this task, we propose a self-localization approach that matches two descriptions of the "road": the one from the geometry of the motion trajectories of cumulatively observed vehicles, and the other one from the aerial laser scan.
An Iterative Closest Point (ICP) algorithm is used to register the motion trajectory in the road section of the laser scan.
The resulting estimate of the transformation matrix represents the sensor pose in a global reference frame. 
We evaluate the results and show that it outperforms other map-based radar localization methods, especially for the orientation estimation.
Beyond the localization result, we project radar sensor data onto a city-scale laser scan and generate a scalable occupancy heat map as a traffic analysis tool.
This is demonstrated using two radar sensors monitoring an urban area in the real world.
\end{abstract}

\section{Introduction}
The transportation of the future will be supported by Intelligent Transportation Systems (ITS). 
ITS consist of roadside sensors with different data modalities and the ability to communicate their understanding of the scene to road users.
Their constant presence and elevated mounting position can assist vehicles with perception tasks, contributing to safer and more efficient individual travel.
They also provide statistical traffic analysis to support urban planning.
Among the sensors in ITS systems, 4D automotive radar has great potential for smart city applications due to its relatively simple mechanical design and low manufacturing cost. 
It can provide a 3D point cloud with a radial velocity attribute over a long distance under adverse weather conditions. 
To promote the application of radar-based ITS, more real-world data should be collected and used as training data for the algorithm development.
Localization, in the sense of determining the installed position and orientation of the sensors, plays a fundamental role in further data processing.
In practice, this task is performed only once during the test bed installation phase using Global Navigation Satellite System (GNSS) technology. 
However, it still requires a great deal of effort and the pose could change exposed to constant environmental influences. 
For mobile sensor setups (e.g. \cite{flexsense,infra2go}), which aim at flexible scenario enrichment, it becomes a more frequent workload. 

\begin{figure*}[t]
      \centering
      \includegraphics[scale=0.55]{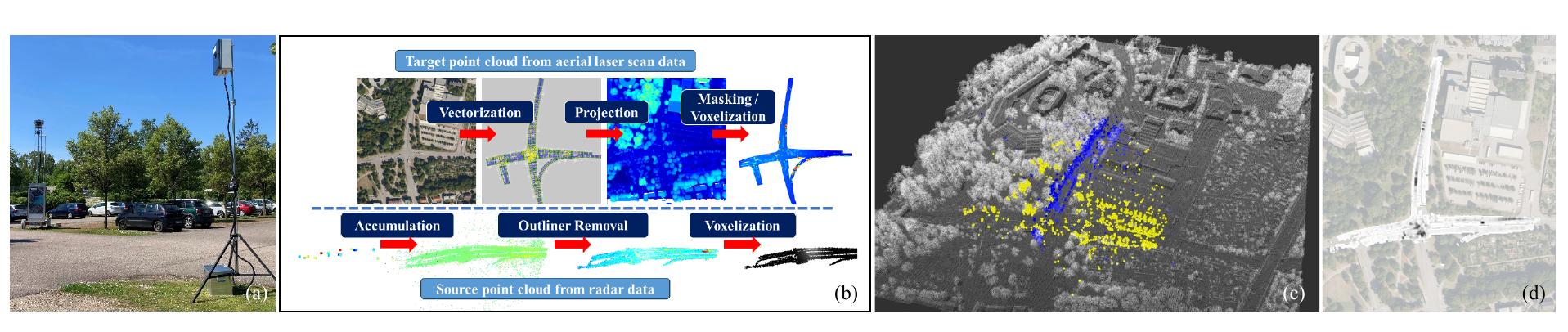}
      \caption{Overview of the scalable radar-based ITS. (a) Flexibly deployable sensor setups. 
      (b) Self-localization based on sensor data and aerial laser scan data. 
      The upper part of the figure illustrates the processing steps to extract the road section from the aerial laser scan with the help of a vectorized map.
The lower part of the figure illustrates the description of the road section using the radar point cloud data.
The voxelized road section from the aerial laser scan and the accumulated radar data are the input to the ICP registration algorithm.
(c) Projection of the radar sensor data onto city scale laser scan.
(d) Occupancy heat map as a tool for traffic analysis. Darker areas indicate high utilisation of the lanelet sector.
      }
      \label{Figure:01_Concept}
   \end{figure*}

In this work, we investigate scalable radar-based roadside perception (Fig. \ref{Figure:01_Concept}) by addressing the localization problem of ITS radar sensors following by a occupancy heat map generated in a sensor network.
Specifically, our localization uses the sensor data directly, without any additional calibration effort in the sensor installation phase. 
To support this process, city-scale aerial laser scan point clouds are used, which are usually available from local authorities.  
The pose of the sensor in a global reference frame can be found by registering the 4D radar sensor data in the aerial laser scan.  
In other words, two descriptions of the concept of "road" are matched:
One from aerial laser scans, 
the other one from the motion trajectories of cumulatively observed vehicles.
This approach runs online.
The result can be checked and corrected on a regular base. 
Based on this localization result, a further contribution of this work is traffic analysis in the form of occupancy heat maps using multiple sensors for a scalable area.
This involves the occupancy of finely subdivided lanelets, featuring both fine-grained and tractable analysis. 
We demonstrate the flexible deployment of radar-based ITS systems on the road side and online generated occupancy heat maps in a scalable manner in the real world.

Our contribution can be summarized as follows:
\begin{enumerate}
    \item A high-precision self-localization method for radar sensors in ITS roadside perception.
It can be run online, replacing the need for additional calibration during the installation phase, and can update the result regularly. 
\item  A demonstration of the use of  radar-based ITS to generate an occupancy head map as a traffic analysis.
The scalability of this analysis is also shown.
\end{enumerate}

The remainder of this paper is organized as follows:
In the second section, we review the related work in sensor localization and traffic analysis visualization.
The third section is dedicated to the localization method. 
Then in the fourth section, we provide the evaluation of our localization method.
The fifth section presents the occupancy heat map using two sensors in a real-world application.
The last section concludes the paper and presents ideas for future research. 

\section{Related Works}
\subsection{Localization of sensors}
With the growing popularity of radar sensors, many localization works have been proposed (e.g. \cite{RadarLoc,KidnappedRadar}). 
The localization problem is closely related to the mapping process and can therefore be divided into map-based approaches and Simultaneous Localization And Mapping (SLAM).
Since roadside sensors in ITS are usually static and have a very close relationship to Geographical Information Systems (GIS), map-based approaches are well suited. 
Open Street Map (OSM) is a community-supported, free world map project. 
Integrating OSM into the localization task has been a popular research direction.
Pioneering works such as \cite{MapLocalization} and \cite{OSMLaser} propose to use OSM for camera and lidar localization.
Hong et al. combine road routing graph and semantic features in OSM to improve the localization accuracy of radar-only systems \cite{radarosm}. 
On a macro scale, OSM provides accurate data for routing and associated localization.
However, if the micro-geometry accuracy of the local map within the view of the sensors is limited, which is usually the case, OSM is not suitable for ITS localization.

The authors of the \cite{RSL-Net,overhead} propose the use of aerial imagery for the localization task of range-finding sensors, both radar and lidar. 
The overhead images, which show the perfect geometry of the road, are now widely available from satellites or from survey flights. 
These images are used to learn radar images along with their rotation and translation. 
The aerial laser scans are also increasingly being provided by local authorities, enabling  their use for the localization problem.
While this hasn't been seen yet, the use of lidar data for radar localization is being explored intensively .
\cite{radaronlidar} learns fake lidar point clouds from radar data for localization. 
A non-learning approach \cite{RoLm} is proposed to localize the radar-based odometry to a lidar map by finding and matching corresponding locations in both modalities. 
Most of the works above address localization of a 2D radar sensor on a moving base. 
An approach for the localization of 4D radar in ITS systems is yet to be seen.  

As we use aerial laser scan along with maps for the localization task, the power of HD maps is harnessed for high accuracy.
Beyond the standard OpenDrive format, we further exploit lane level information.
Lanelet2 \cite{lanelet2o,lanelet2p}, developed by Poggenhans et al., is an open-source library and framework for detailed high-precision maps based on the concept of Lanelet \cite{lanelet1}.
A lanelet is the smallest unit within which traffic rules do not change. 
In addition, the topology formed by the lanelets is constant. 
As a comprehensive and flexible map format, Lanelet/Lanelet2 has recently been widely adopted in ITS tasks (e.g. \cite{GOHOME}-\cite{DRG}).
Another example is the CommonRoad project \cite{ComRoad}, which aims to provide various standardized motion planning scenarios, also uses a lanelet-based map format. 
It further provides a powerful toolkit \cite{ComRoadSceDe} that allows easy conversion between many different map formats. 

\subsection{Traffic Analysis and ITS Data Visualizations}
The long-range sensing capability of radar enables powerful traffic analysis with wide coverage at low sensor count and cost. 
However, city scale traffic analysis requires methods and tools to efficiently fuse the data and generated insights for downsteam tasks \cite{visualization}. 
The line chart, the bar chart, the heat map, and the geospatial map are the popular techniques for visualization \cite{intracity}.
The authors in \cite{spatiotemporal} use heat maps for traffic analysis, focusing on spatiotemporal aggregation and distillation.
In \cite{twatcher}, regional fingerprinting based on heat maps is introduced for monitoring traffic. 
Another example is VAIT \cite{VAIT}, where the heat map is used for visual analysis of urban transportation.
We adopt the use of the heat map for occupancy in the current situation as a basis for further traffic analysis.

\section{Localization}
\label{Localization}

The use of a GNSS-based localization approach for ITS systems has considerable limitations. 
Due to the clearance profile of road design, urban ITS sensors usually have to be installed near trees and buildings, so the multipath and non-line-of-sight (NLOS) effects could negatively affect the accuracy of the GNSS system.
In addition, the deployment of a flexible sensor system should also take into account the influence of cloudy or rainy weather, where the GNSS receiver may not be able to obtain reasonable readings.
The radar sensor data, on the other hand, shows strong geometric features in the long term due to the fixed road structure. 
After a sufficiently long  sampling period, the vehicles observed could pass through any accessible point in the scene.
Furthermore, the sensor data is available regardless of weather conditions.

\subsection{Method Overview}
The approach is briefly illustrated in Fig. \ref{Figure:01_Concept}(b).
The data flows through two pre-processing pipelines and is then fed to the ICP algorithm \cite{icp}. 
In the upper part of the Fig. \ref{Figure:01_Concept}(b), the aerial image is vectorized to obtain the HD map. 
This step is currently conducted manually, but could potentially be extended to an automatic process similar to \cite{lanedetection}.
It is important to preserve the geometric features of the road, we therefore use the Lanelet2 \cite{lanelet2o} format for lane-level detail.
The map is then projected onto the laser scan frame to mask the portion of the road from the entire scan.
The point cloud of the road then goes through the outliner elimination process (e.g. DBSCAN \cite{dbscan}) to remove noise such as reflections from trees.
We then voxelize the point cloud into 0.5m $\times$ 0.5m grids.
The grid center points are used.
This completes the generation of the target point clouds in the ICP algorithm.
For real use cases, data from local authorities (e.g. \cite{LaserScan}) can be used.
In simulations, road meshes can be converted to point clouds.

The second part of the pre-processing is the generation of motion trajectories from radar sensors representing the road section.
The first step is to accumulate radar sweeps. 
A sufficient number of frames $n_f$ should be chosen.
We set $n_f$ to 2000.
Since the radar works with 20Hz, we could check and correct the pose of the radar, using the data from 100s.
Given that the radar could provide about 500 points per scan, a total of 1 million points could be collected, which is too much for the registration algorithm to run efficiently.
Since our goal is to retrieve the points that represent the road and the points there should be moving in the long run, static reflections are filtered out.
This is done using the Doppler measurement, only points with a radial speed above $0.15m/s$ are considered. 
A statistical analysis shows that more than $93\%$ of the points have a speed below this threshold, making this criterion an efficient filter that could later speed up the ICP process.
The point cloud then goes through an outliner removal process. 
Ghost reflections and small clusters of points are removed using DBCAN.
With the proper parameters, the largest cluster represents the points on the road. 
The cluster is then voxelized as source point clouds for the ICP algorithm.
This pipeline is shown in the lower part of the Fig. \ref{Figure:01_Concept}(b).

\subsection{Localization with ICP}
The ICP algorithm establishes the data association between the source and target point clouds via nearest neighbour, and reduces the distance between them to obtain the transformation matrix \cite{icp}. 
\begin{equation}\label{equation:icp}
\mathbf{T} = \underset{\mathbf{T}}{\text{arg min}} \, \sum^{n} \|\mathbf{T} \mathbf{p}_r - \mathbf{p}_{als}\|^2.
\end{equation}
As shown in Eq. \ref{equation:icp}, the associated $n$ pairs of the source point $\mathbf{p}_r \in \mathcal{P_{\text{r}}}$ from the radar and the target point $\mathbf{p}_{als} \in \mathcal{P_{\text{als}}}$ from the aerial laser scan go through an optimization step to find the transformation $\mathbf{T} \in {SE(3)}$ that gives the minimum sum of distances $\|\mathbf{T} \mathbf{p}_r - \mathbf{p}_{als}\|^2$. 
In our implementation, it follows a multi-scale approach and is executed multiple times.
To initialize the ICP algorithm, we make a manual coarse selection of the deployment position, such as a point on the map, and the estimated orientation, such as "east", "north", etc.
The first ICP process is executed with a large allowed correspondence distance.
The following ICP runs use a reduced correspondence distance of 2$\times$ and 1$\times$ voxel size.


For continuous computation, a Kalman Filter \cite{kf} with a static motion model is implemented. 
The state vector $\bm{x}$ consists of the position $\bm{p}$ as a 3D coordinate in the UTM (Universal Transverse Mercator) frame with a predefined origin and a quaternion $\bm{q}$ for the orientation.
\begin{equation}
\bm{x} = [\bm{p}, \bm{q}]^T = [x_p, y_p, z_p, x_q, y_q, z_q, w_q]^T.\end{equation}

The setup presumes a static model.
The prediction can be simplified as follows
\begin{equation}\bm{\hat{x}}_{k+1,k} = \bm{\hat{x}}_{k,k} + \bm{w}_k, \bm{w}_k \sim \mathcal{N}(\bm{0}, \bm{Q}_k), \end{equation}

\begin{equation}\bm{P}_{k+1,k} = \bm{P}_{k,k} + \bm{Q}_k,\end{equation}
where
$\bm{Q}_k$ is the uncertainty of the static model. 
We then use the outcome from the ICP algorithm to correct the prediction to obtain the estimate.
Since ICP generates all values in the state vector, we can simplify the measurement model to
$\bm{z}_k = \bm{x}_k +\bm{v}_k, \bm{v}_k \sim \mathcal{N}(\bm{0}, \bm{R}_k)$.
The correction procedure is given by
\begin{equation}\bm{K}_k = \bm{P}_{k,k-1}(\bm{P}_{k,k-1} + \bm{R}_k)^{-1},\end{equation}
\begin{equation}\hat{\bm{x}}_{k,k} = \hat{\bm{x}}_{k,k-1} + \bm{K}_k(\bm{z}_k-\hat{\bm{x}}_{k,k-1}),\end{equation}
\begin{equation}\bm{P}_{k,k} = (\bm{I} - \bm{K}_k)\bm{P}_{k,k-1}(\bm{I} - \bm{K}_k)^T + \bm{K}_k\bm{R}_k\bm{K}_k^T,\end{equation}
where
$\bm{R}_k$ is the measurement uncertainty.
The Kalman Filter works in conjunction with the ICP on a 5-second cycle. 
For each run, radar point clouds are collected within a rolling window spanning the last 2000 frames (equivalent to approximately 100s at a sampling rate of 20Hz). 
These gathered points are then used to generate vehicle traces for the localization.
This method filters out transient changes in sensor pose while preserving long-term changes, which are caused by environmental factors such as wind, rain drops, and heat.
   
\section{Evaluation of Localization}
\begin{figure*}[t]
      \centering
      \includegraphics[scale=0.5]{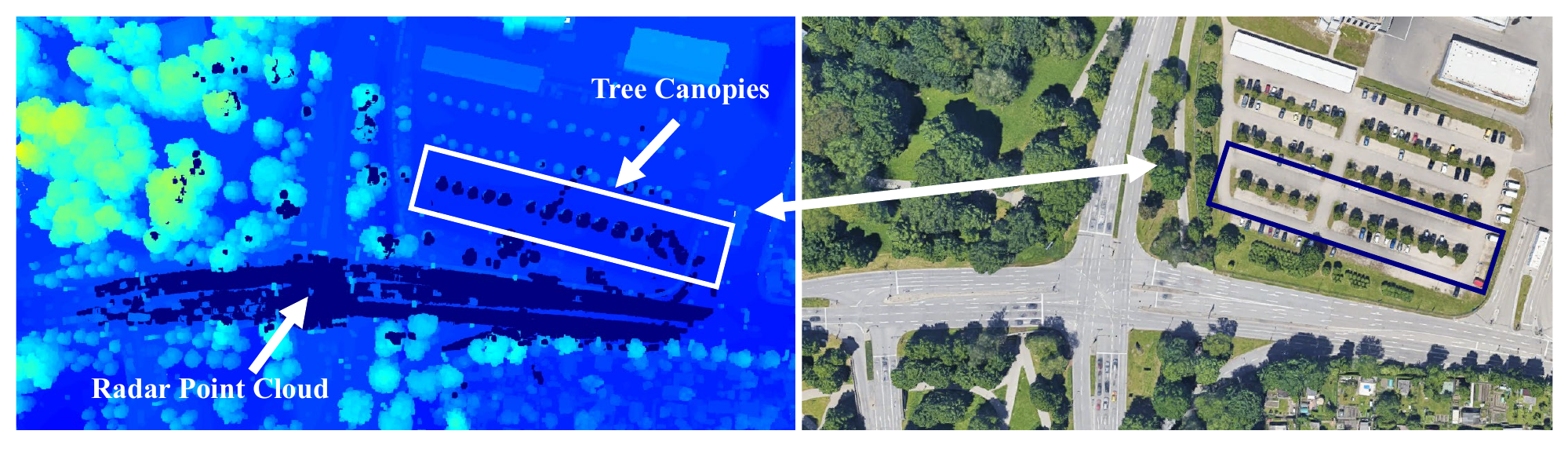}
      \caption{The qualitative evaluation of the localization result is shown in the left figure.     
      Dark blue points are the radar point clouds.
      Blue to green points are aerial laser scan points.
The aligned dark blue clusters exactly overlap the tree canopy (marked by the rectangles).
      The right image from Google Earth is used as a reference for the reader to understand the scene.
      }
      \label{Figure:02_Qualitative}
   \end{figure*}
We evaluate the localization algorithm using both real world data and simulation data.
\subsection{Localization of Radar Sensors in Real World}
For real-world localization, radar data is gathered in Ingolstadt, Germany with a mobile sensor system \cite{flexsense}.
It contains a Continental ARS548 long-range radar sensor and a Nvidia Orin embedded system to collect and process the data.
The radar has an azimuth range of 120° and an elevation range of 30° and operates at 20Hz.
Each frame provides about 500 detections in a point cloud format.
The aerial laser scan is retrieved from open data provided by the Agency for Digitisation,
High-Speed Internet and Surveying of Bavaria \cite{LaserScan}.

A qualitative assessment is shown in Fig. \ref{Figure:02_Qualitative}. 
The localization accuracy is demonstrated using the tree canopies as a reference in a parking lot.
The radar data was gathered from 23,950 frames over a period of approximately 20 minutes on a windy day. 
The point clouds are first passed through a velocity filter.
Only points with a speed greater than $0.15m/s$ are retained.
The cloud is then filtered using the DBSCAN algorithm (with $\epsilon = 0.5, n_{min} = 10$). 
The points on the canopy represent the reflection from the rustling leaves.
It can be seen that the points are grouped in clusters with a clear separation between them.
The clusters overlap very well with the laser scan point clouds of the trees.
This shows that the algorithm is accurate in determining position and orientation.

\begin{figure}[H]
      \centering
      \includegraphics[scale=0.28]{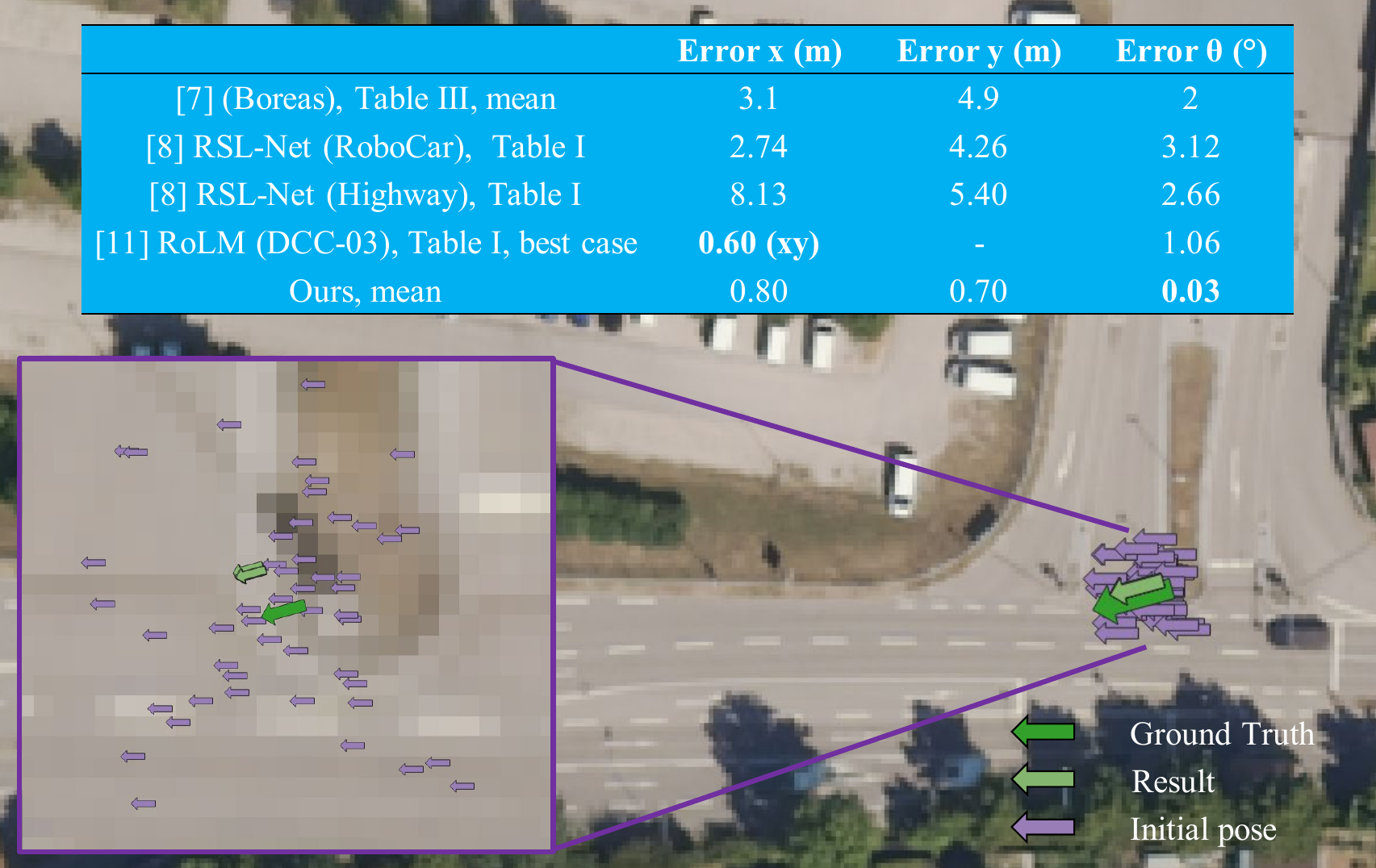}
      \caption{Quantitative evaluation of the localization.
      50 initial poses are randomly selected around the ground truth value of the sensor setup for the ICP process.
      The error between the ICP result and the ground truth is given in the table.}
      \label{Figure:03_Quantitative_point}
   \end{figure}

\begin{figure}[h]
      \centering
      \includegraphics[scale=0.42]{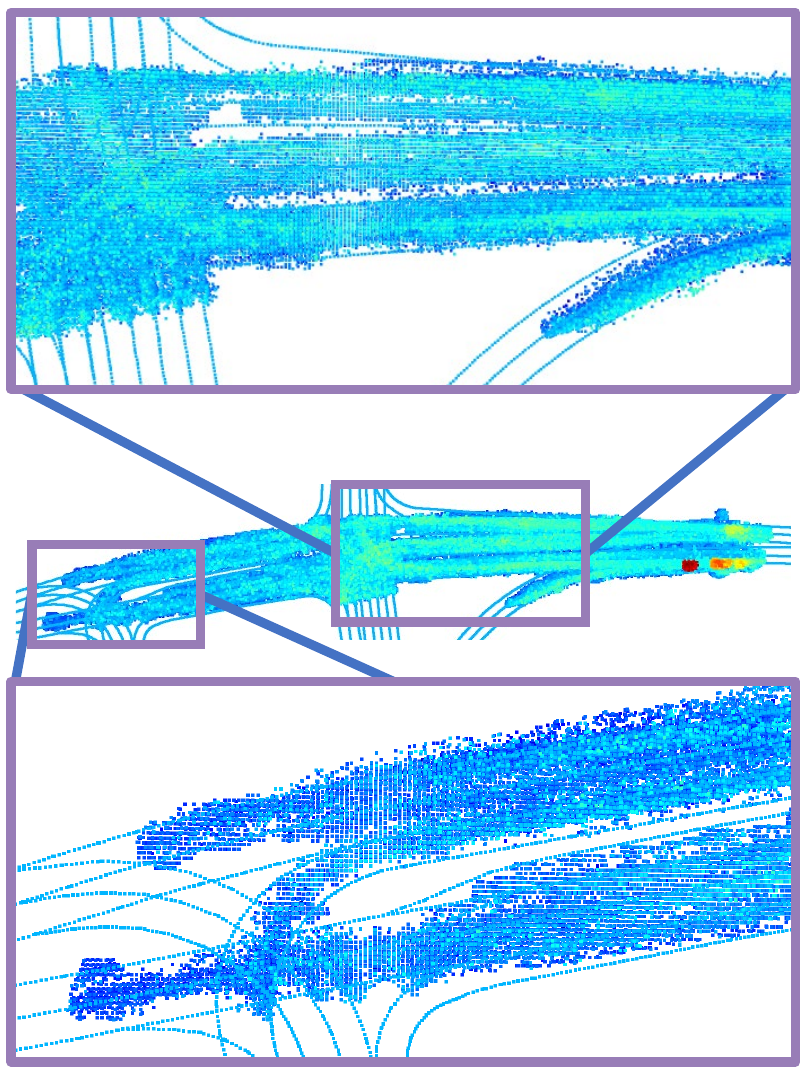}
      \caption{Alignment of the radar point cloud to the road.
      The lane description is  manually created based on an aerial image (dashed lines).
      The lane information is also used for the masking process in Fig. \ref{Figure:01_Concept}(b).
The upper zoomed image shows that the green area on the road corresponds to the untraveled area in the trajectory point cloud.
The lower image shows that the traces of the left-turning vehicles match the lane description.}
      \label{Figure:04_Quantitative_lane}
   \end{figure}

For a quantitative evaluation, we place the sensor setup at a random position around an intersection to gather data for localization.
We then compare the results with a ground truth determined with a GNSS device (ArduSimple 2B) using RTK (real-time kinematics) correction.
The accuracy of the GNSS measurement is $2 cm$.
50 positions in the map surrounding the test spot are generated as initial values for the ICP algorithm.
The actual sensor location, the results, and the initial starting points are marked in the Fig. \ref{Figure:03_Quantitative_point}.
The 2D errors are also shown.
The test runs have similar height results, with an average error of 0.59m.
The average 2D position error is 1.06m, which is comparable to other map-based approaches \cite{radarosm,RSL-Net,RoLm}, some results of which are also shown in the figure along with the used dataset.
The resulted pose are densely distributed around a point, indicating convergence.
It also indicates a hidden pattern in the point cloud data.
The height error may be due to changes in terrain elevation.
The yaw error is extremely low compared to the other methods. 
This is very important for ITS roadside perception, as we are aiming for long-range detections (up to 300m) with a single sensor, large angular deviations will cause the detection to drift completely away from its true position. 
In Fig. \ref{Figure:04_Quantitative_lane} we see that the vehicle's tracks are aligned with the driving lane, while also leaving the green area, road divider, and other areas unoccupied (upper part).
It further depicts that the left-turning vehicles stay correctly in the lane (lower part). 
This indicates an acceptable accuracy for traffic analysis.

\subsection{Localization of Radar Sensors in Carla Simulation}
\begin{figure}[h]
      \centering
      \includegraphics[scale=0.3]{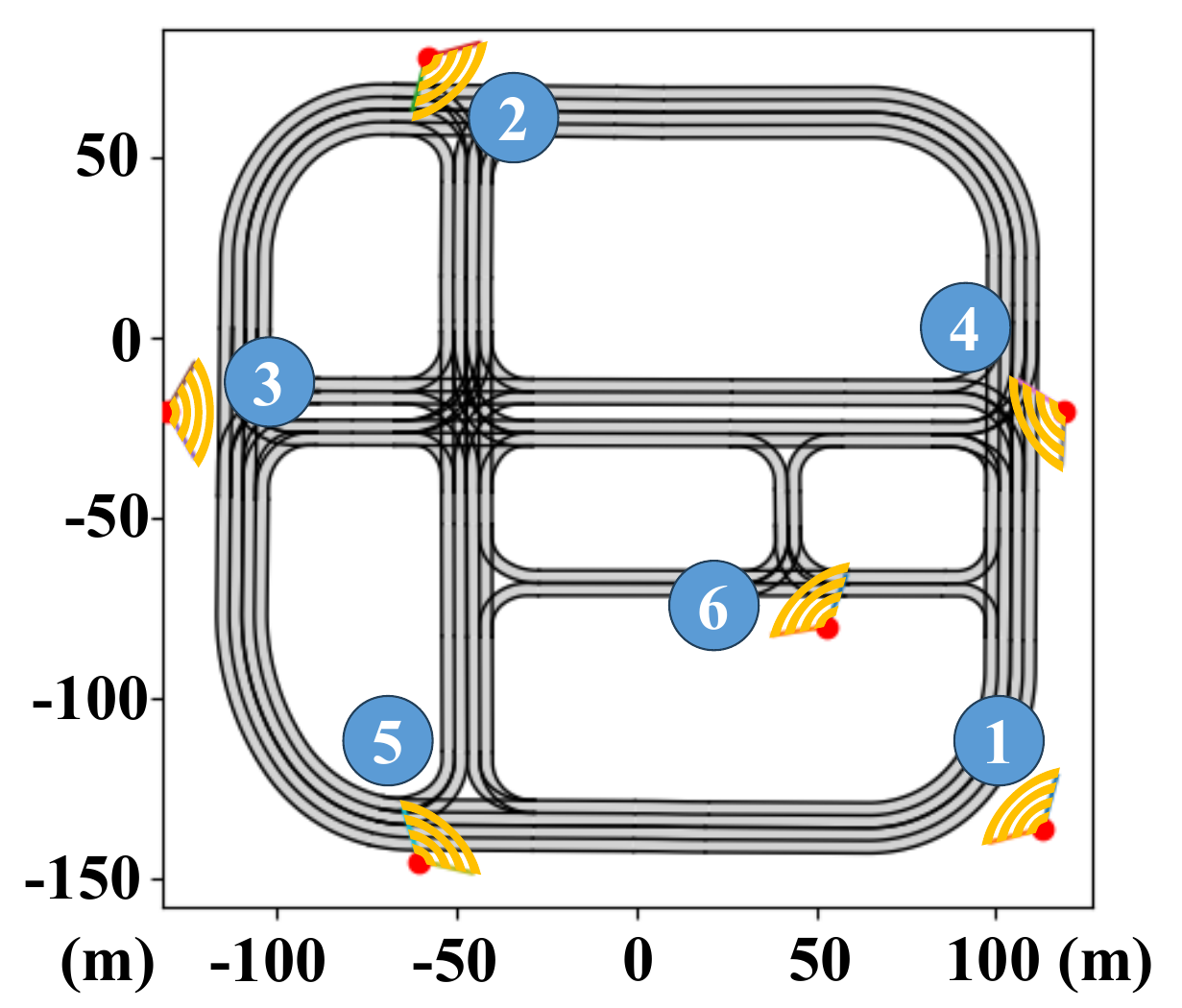}
      \caption{Test points in the Carla map Town 10.
      The points and the line segments show the position and orientation of the radar sensors in test DY.
      In test DO the position is maintained but the orientation is more flexible and not shown in the figure.}
      \label{Figure:05_Carla}
   \end{figure}

We evaluate the localization method in Carla simulator's Town 10 map.
A laser scan is not directly available for this map.
Instead, we convert the road mesh into a point cloud.
We then crop the road section using the HD map of the town to remove the parking lanes.
We run two sets of experiments that change the location and direction of the radar sensors.
In the first set of experiments, we place the radar sensor with \underline{d}ifferent \underline{y}aw angles (DY).
In the second set of tests, the sensors are placed at \underline{d}ifferent positions and have a more flexible \underline{o}rientation (DO).
In both sets of tests, the sensors are placed as shown in Fig. \ref{Figure:05_Carla}
The sensors have a maximum sensing range of 200m.
As with the real-world radar, the azimuth angle of the sensors ranges from -60° to 60°. 
Their orientations in the first set are also shown in Fig. \ref{Figure:05_Carla}.
The elevation range is 30°.
Table \ref{table: carla} shows the results of the experiments.
The 2D position error is less than 1 meter, except for R2 and R6 in DO.
The altitude estimate is always off by about 1.06 meters on average.
The average yaw errors are 0.38° (DY) and 0.75° (DO).
In the best case scenario, it could only be off by 0.02° (R1 in DY).
The roll and pitch errors in DO are 0.28° and 0.33° respectively. 
This also outperforms other map-based methods \cite{radarosm,RSL-Net,RoLm}.
We believe that the favorable result of Carla's 2D position and orientation is due to its complex observable geometry and simple road geometry.
The evaluation demonstrates the potential of our approach.
\begin{table}[h]
\caption{Result of tests in carla}
\label{table: carla}
\begin{center}
\begin{tabular}{|c|c|c|c|c|c|c|}
\hline
DY Error&  X(m) & Y(m) & Z(m) &  yaw(°)\\
\hline
R1& -0.71 & 0.68 & -1.02 &  0.02\\
R2& -0.79 & -0.67 & -1.05 & 0.06\\
R3& 0.59& -0.69 & -1.05 &  0.11\\
R4& -0.09 & -0.10 &-1.06 &  0.30\\
R5& -0.27 & 0.03 & -1.03 & 0.25\\
R6& -0.95 &-0.69 & -1.13 & 1.53\\
\hline

\end{tabular}

\vspace*{0.1 cm}

\begin{tabular}{|c|c|c|c|c|c|c|}
\hline
DO Error & XY(m) & roll(°) & pitch(°) & yaw(°)\\
\hline
R1& 0.50 & 0.21 & 0.12 & 0.58 \\
R2& 5.06 & 0.16 & 0.29 & 0.06\\
R3& 0.59&0.15 & 0.24 &  0.16\\
R4& 0.19 & 0.64 & 0.41 &  0.34\\
R5& 0.20 & 0.19 & 0.49 & 0.95\\
R6& 3.35 & 0.31 & 0.41 & 2.42\\
\hline
\end{tabular}
\end{center}
\end{table}

\section{Scalable Dynamic Occupancy Map}
   \begin{figure*}[t]
      \centering
      \includegraphics[scale=0.5]{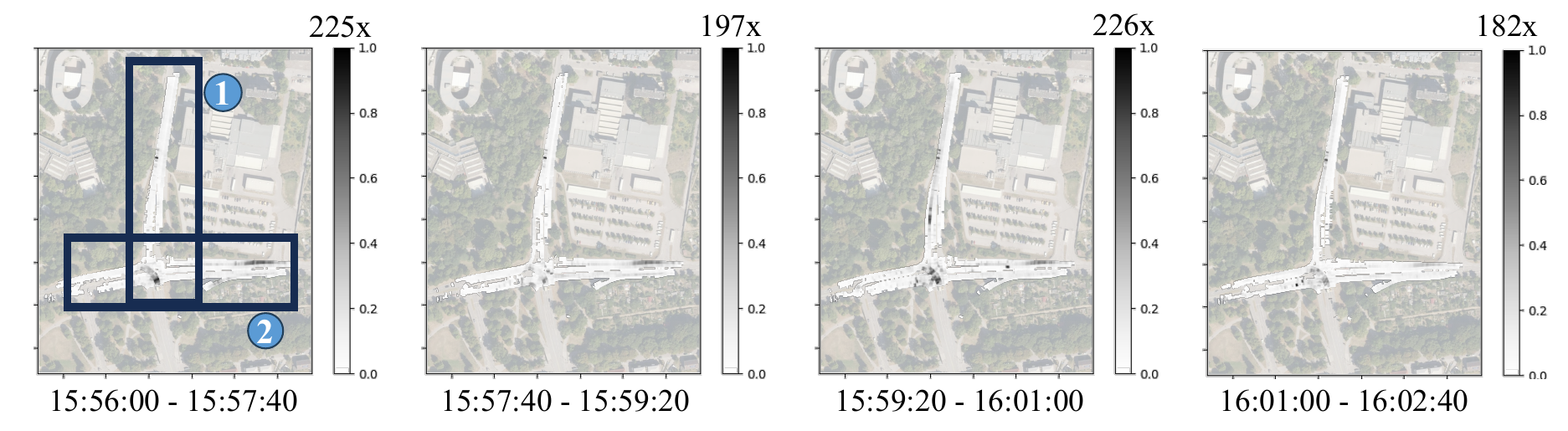}
      \caption{The global occupancy heat map as a tool for traffic analysis.
      The heat map is created using data from two sensors.
      Sensor 1 monitors the road section from top to bottom.
      Sensor 2 monitors the road section from right to left.
      The maximum range is about 300 meters.
      The intersection can be monitored by both sensors.
      The time difference is indicated below the figures.
The dark color indicates a high occupancy.
      The number above the color scale is the maximum occupancy of the polygon nodes in the time scope.      
      For example, 225 implies an occupancy of 11.25s out of 100s ($225 \times 50 $ms).
      }
      \label{Figure:09_OccupancyMap}
   \end{figure*}
After locating multiple radar-based ITS setups in the city (Fig. \ref{Figure:01_Concept}(c)), large-scale traffic analysis could be conducted. 
It performs better than sensors installed at specific points, such as inductive circuits or ultrasonic detectors, within the area. 
Compared to vision-based systems, the radar-based systems generate less data traffic and can monitor larger areas with fewer sensors.
Using an occupancy heat map as a tool, we can display the current traffic status and perform spatial statistical analysis. 
\subsection{Using the HD Map as Scaffold}
We design the occupancy heat map based on a sublane level HD map. 
We define the urban occupancy heat map as $M=\{P\}$, where $P$ are the nodes of the locations.
Unlike the voxelized scene representation, each node in this map is defined as a sub-lane level tile of the road. 
It is a four-vertex polygon with a width
of approximately 3m, equal to the width of the lane and a length of about 0.5m, close to the standard step of a pedestrian. 
These shapes come from an HD map of lanes in Lanelet2 (OSM) format. 
First, we interpolate the left and right edges of the lanelet to get finer points.
Then, every four points that enclose a smallest rectangle form a finer lanelet and are stored as a polygon.
A unique ID is assigned to these polygon nodes as a reference. 
Similarly to lanelets at junctions, it is possible to have overlapping nodes.

\subsection{Local Sensor Data Processing}
To analyze the occupancy of the scene, we first project the sensor data onto the polygons.
Whenever the radar sensor delivers a point cloud, we check its footprints on the east-north plane and assign them to the corresponding polygons.
We then add the current timestamp to the polygon node's "occupancy time" attribute, indicating that the node is occupied at that timestamp.
As there can be a very large number of potentially occupied polygons, the R-tree spatial indexing \cite{rtree} is used for efficient assignment.
This compresses the local sensor data into a list of indices of occupied polygons, which can greatly reduce the data volume.
Note that the data can be filtered based on the Doppler measurement, the road clearance profile or the radar cross section measurement.

\subsection{Global Occupancy Heat Map}
\begin{figure}[H]
      \centering
      \includegraphics[scale=0.25]{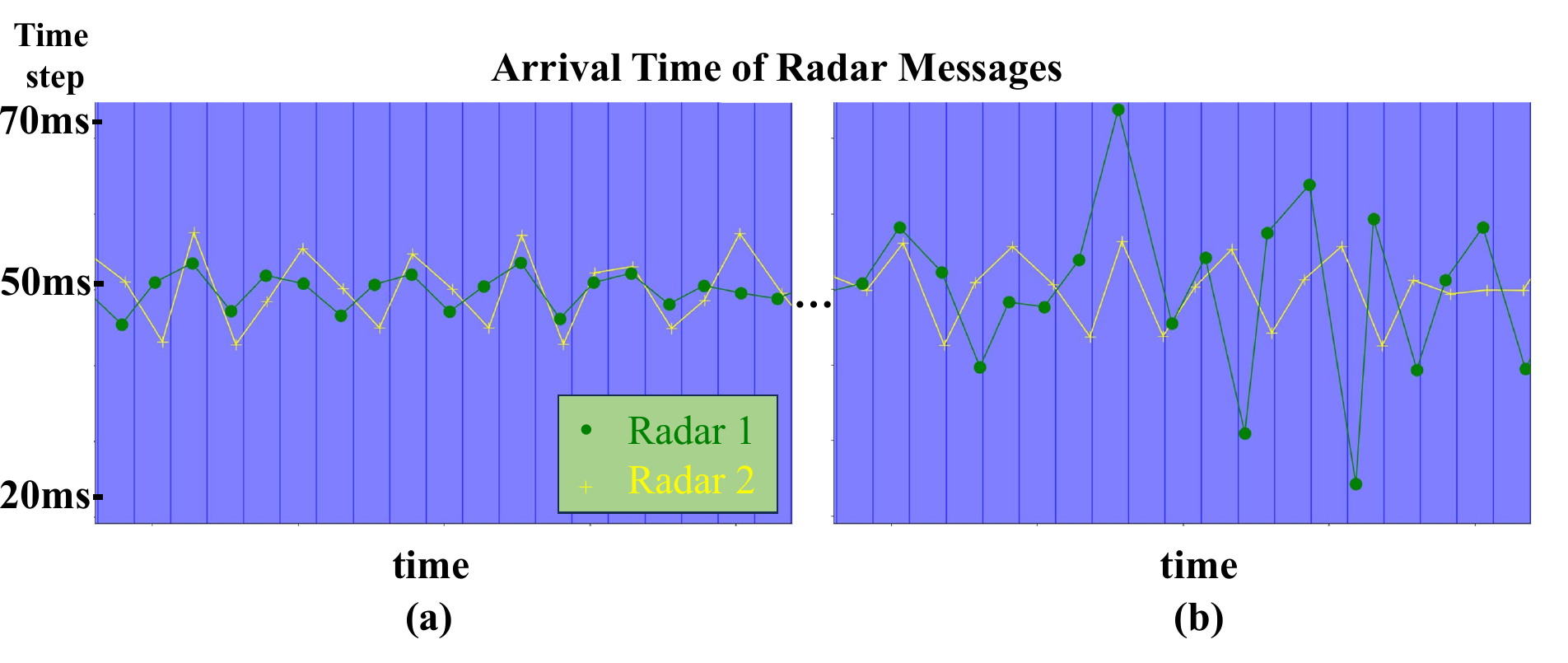}
      \caption{A diagram of message arrival time and time synchronization in the occupancy map. 
      The X-axis shows the time stamp when messages arrive.
      The Y-axis shows the time step between two adjacent messages:
      $t_n = t_{n-1} + y$. 
      The blue rectangles represent sampling windows of 50ms.}
      \label{Figure:08_Sync}
   \end{figure}

After connecting the sensors to a network, we can create a global occupancy heat map by merging all the data frames into a single scene representation.
We first use NTP (Network Time Protocol) to synchronize the time of all data processing units.
A 5G module is used to connect the sensor and its data processing unit to the internet. 
The sensors operate at a approximately constant frequency.
However, fusing sensor data into an integrated map is not as trivial as adding up all the lists of occupied polygons and time stamping the occupancy.
Sensor coverage areas sometimes overlap, creating redundancy. 
As a result, the overlapping areas receive more frequent observations and therefore  exhibit a higher occupancy. 
This will incorrectly suggest more occupants in those areas.
To prevent problems, it is necessary to establish a uniform update rate for the global occupancy map that matches the radar update rate.
Depending on whether the global occupancy map is generated in a field data processing unit or in the cloud, the update rate can either adopt the frame rate of the connected sensor or be defined independently.
Despite efforts to synchronize the clocks in the data processing units, there is a noticeable time offset in the sensor data depending on their starting time (see Fig. \ref{Figure:08_Sync}).
The data is updated locally at a frequency of approximately 20Hz, but there may be variations due to the time taken by the on-board algorithms in the sensor.
The data for generating the global occupancy heat map is then collected within each 50ms time window.
Fig. \ref{Figure:08_Sync}(a) illustrates that if the frame rate is close to 20Hz, we can collect the data and build an occupancy map within each sampling window.
If the data frequency deviates significantly (see Fig. \ref{Figure:08_Sync}(b)), two messages may arrive during the same sampling period, resulting in both messages being added to that time step.
If no message is received during a particular time step, it should be left empty.

The Occupancy Heat Map can be generated either instantaneously to show the current state of road occupancy or cumulatively by counting the reported occupancy time stamps to show the statistical state of congestion.
The volume of data can be controlled by the number of polygons.
Fig. \ref{Figure:09_OccupancyMap} shows an example where two sensors are deployed in the city and localized using the proposed method.
The traffic changes at the intersection are shown with the changes in the occupancy status,  from left to right.
On the right side of the map, congestion can be observed due to an increased volume of traffic in the right turn lane.
Similarly, the left-turn queuing area at the intersection from below is congested, indicating a high volume of traffic moving from the bottom to the left part of the area.

\section{Conclusion}
In this work, we investigate 4D radar-based ITS to leverage the benefits of long sensing range under diverse weather conditions.
A method for self-localization is proposed. 
It exploits the idea of matching the concept of "road" in two data modalities: aerial laser scan and live radar sensor data.
On the one hand, road sections are represented by a point cloud from an aerial laser scan, which is extracted with the help of aerial images and HD maps. 
On the other hand, the radar point cloud of motion trajectories generated by the observed vehicles also describes the geometry of the road sections.
These two point clouds are matched using the ICP algorithm. 
The output transformation matrix delivers solutions to the localization task. 
An additional Kalman Filter is integrated into the localization process, allowing corrections during deployments.
Our results outperform those of other map-based radar localization works, especially for the orientation.
Using the localization results, we demonstrate the creation of global occupancy heat maps by synchronizing networked radar-based sensor setups for scalable traffic analysis.
Based on this work, we contribute to the flexible deployment of radar-based ITS for traffic analysis.
Future work will focus on detecting individual moving objects on the road and matching their trajectories to the corresponding lane to furthur improve the localization accuracy.
Another extension is to add object information to the occupancy heat map for further traffic analysis.
\bibliographystyle{ieeetr} 
\bibliography{refs} 

\end{document}